\documentclass[journal,twoside,web]{ieeecolor}
\usepackage[table]{xcolor}
\usepackage{generic}
\usepackage{cite}
\usepackage{amsmath,amssymb,amsfonts}
\usepackage{algorithmic}
\usepackage{graphicx}
\usepackage{textcomp}
\usepackage{wrapfig}
\usepackage{algorithm,algorithmic}
\usepackage{hyperref}

\hypersetup{hidelinks=true}
\usepackage{textcomp}
\usepackage{booktabs} 
\usepackage{caption}
\usepackage{makecell} 
\usepackage{longtable}
\usepackage{array}
\usepackage{changes}
\usepackage{xcolor}
\usepackage[switch]{lineno}
\providecommand{\refname}{References}

\def\BibTeX{{\rm B\kern-.05em{\sc i\kern-.025em b}\kern-.08em
    T\kern-.1667em\lower.7ex\hbox{E}\kern-.125emX}}
\begin{document}
\title{Deep Learning-Based Estimation of Ground Reaction Forces in Parkinsonian Gait Using an Optimized Set of IMU Data}
\author{Run Lin, \IEEEmembership{Member, IEEE}, Yingtian Tang, Jiawen Xu, Dongfei Huo, Lefan Wang, \IEEEmembership{Member, IEEE}, Helen Dawes, Dominic J. Farris, Dong Wang, and Xijin Hua, \IEEEmembership{Member, IEEE}
\thanks{This work was supported by the Royal Society, UK [grant number RG\textbackslash R1\textbackslash 251639 and IES\textbackslash R3\textbackslash 243317], China Scholarship Council and University of Exeter PhD Scholarships.}
\thanks{Run Lin is with the Department of Engineering, Faculty of Environment, Science and Economy, University of Exeter, Exeter, EX4 4QF, UK (e-mail: rl696@exeter.ac.uk).}
\thanks{Yingtian Tang is with the School of Computer and Communication Sciences, École Polytechnique Fédérale de Lausanne (EPFL), CH-1015 Lausanne, Switzerland (e-mail: yingtian.tang@epfl.ch).}
\thanks{Jiawen Xu is with the Department of Engineering, Faculty of Environment, Science and Economy, University of Exeter, Exeter, EX4 4QF, UK (e-mail: jx346@exeter.ac.uk).}
\thanks{Dongfei Huo is with the Department of Engineering, Faculty of Environment, Science and Economy, University of Exeter, Exeter, EX4 4QF, UK (e-mail: dh691@exeter.ac.uk).}
\thanks{Lefan Wang is with Institute for Manufacturing, Department of Engineering, University of Cambridge, Cambridge CB3 0FS, UK (e-mail: lw683@cam.ac.uk).}
\thanks{Helen Dawes is with the NIHR Exeter BRC, University of Exeter Medical School, Exeter, EX1 2LU, UK (email: H.Dawes@exeter.ac.uk).}
\thanks{Dominic J. Farris is with the Department of Public Health and Sport Sciences, Faculty of Health and Life Sciences, University of Exeter, St Luke's Campus, Exeter, EX1 2LU, UK (e-mail: D.Farris@exeter.ac.uk).}
\thanks{Dong Wang is with the Department of Engineering, Faculty of Environment, Science and Economy, University of Exeter, Exeter, EX4 4QF, UK (e-mail: d.wang2@exeter.ac.uk).}
\thanks{Xijin Hua is with the Department of Engineering, Faculty of Environment, Science and Economy, University of Exeter, Exeter, EX4 4QF, UK (e-mail: x.hua@exeter.ac.uk).}}

\maketitle

\begin{abstract}
Accurate gait analysis in Parkinson's disease (PD) typically relies on laboratory-based systems to capture biomechanical data, such as ground reaction forces (GRFs). Estimating GRFs using inertial measurement units (IMUs) provides a feasible alternative. However, this approach remains challenging in pathological gait like PD due to its high variability and complexity. Moreover, existing monitoring approaches often require multiple body-mounted sensors, which limit practicality and reduce patient compliance. To date, no study has investigated the application of deep learning approaches to address this challenge. This study proposes, for the first time, a deep learning framework to estimate bilateral vertical GRFs (vGRFs) in PD using an optimized set of wearable IMUs. A hybrid CNN-BiLSTM model was trained separately on data from 61 PD patients and 65 healthy controls (HC) using 13 IMUs. The model achieved high intra-subject accuracy ($R^2$ = 0.98) and strong inter-subject generalization ($R^2$ = 0.93 for HC, $R^2$ = 0.91 for PD). Sensor configuration was found to significantly influence estimation accuracy, with optimal sensor placement varying between PD patients and HC. For PD patients, estimation accuracy dropped markedly when reducing to a single IMU. The optimal configuration for PD used four IMUs. We identified a minimal setup with only two IMUs still enabled robust estimation. This compact setup offers a practical and scalable solution. Overall, the proposed approach supports the development of wearable vGRF-based gait analysis systems for Parkinsonian gait and potentially other pathological conditions, enabling accessible clinical assessments, remote monitoring, and personalized rehabilitation.
\end{abstract}


\section{Introduction}
\label{sec:introduction}
\IEEEPARstart{B}{Y} 2050, it is estimated that approximately 25.2 million people worldwide will be living with Parkinson's disease (PD), posing significant public health challenges for patients, their families, caregivers, communities, and society \cite{Sue080952}. Monitoring the ground reaction forces (GRFs) during pathological gait in PD provides critical insights for personalized diagnosis, treatment planning, rehabilitation strategies, and long-term disease management. GRFs, particularly vertical ground reaction forces (vGRFs), encapsulate critical clinical and biomechanical markers in PD. Altered vGRFs reflect postural instability, asymmetrical loading, impaired propulsion, and heightened fall risk. Furthermore, vGRF variability has been proposed as a potential biomarker for disease progression and treatment response \cite{RF13,RF14,RF15}. Force plates are widely regarded as the gold standard for vGRF measurement \cite{RF16}. However, they are typically constrained by their dependence on controlled laboratory environments, limiting their feasibility for large-scale clinical applications or continuous, all day monitoring. Beyond force plates, studies have explored GRF measurement using rigid force-sensing shoes and instrumented insoles \cite{RF17,RF18}. Nevertheless, these devices are also limited by their low reusability, substantial weight, and susceptibility to environmental constraints, severely restricting their utility in large-scale, real-world scenarios.

Wearable sensors, particularly inertial measurement units (IMUs), provide a practical option for motion assessment because of their portability, environmental flexibility, and relatively low cost~\cite{RF19}. As summarized in Table \ref{tab:study_summary}, prior studies have used wearable IMU sensors to estimate GRFs. Existing IMU-based GRF estimation approaches can be broadly categorized into physics-based and data-driven methods. Representative physics-based approaches include extended Kalman filters (EKFs)\cite{refai2020portable}, musculoskeletal modelling\cite{eltoukhy2017prediction}, and inverse dynamic models based on Newton-Euler equations~\cite{Liu2023}. Although these approaches are physically interpretable, they commonly rely on predefined kinematic and dynamic assumptions, limiting their adaptability to complex or non-periodic motions. Furthermore, physics-based models present several limitations. First, it is difficult to simultaneously estimate gait dynamics across varying walking speeds and elderly gait, where a more sophisticated model is required. Second, pathological gait, like PD, may violate several fundamental assumptions underlying physics-based models, including assumptions of gait symmetry, steady-state periodicity, and simplified force distribution during double-support phases, which may reduce accuracy and generalizability. Third, the presence of double-support phases during walking introduces additional challenges in separately estimating the GRFs of both limbs\cite{firouzi2025biomechanical}.

To address the limitations inherent in physics-based models, data-driven models, including regression analyses and advanced machine learning algorithms, have been increasingly proposed~\cite{wang2022hybrid}. These methodologies are highly effective in modelling complex motion patterns due to their ability to automatically capture high dimensional non linear relationships and maintain robustness against data noise. Additionally, once adequately trained, they provide rapid inference capabilities, making them highly suitable for estimation tasks. Hossain et al.~\cite{hossain2023estimation} employed three IMUs positioned on the thigh, shank, and foot to develop the Kinetics-FM-DLR-Ensemble-Net model across multiple walking tasks. They reported high Pearson correlation coefficients ranging from 0.923 to 0.929 in healthy individuals. Bach et al.~\cite{bach2022predicting} further demonstrated the effectiveness of machine learning methods by using two IMUs placed on both tibiae during a preferred speed walking task, achieving coefficients between 0.91 and 0.93 in healthy individuals. Collectively, these studies highlight the potential of IMUs as effective solutions for continuous, all-day monitoring of gait and movements in clinical and home-based environments.

\begin{table}[h]
    \centering
    \renewcommand{\arraystretch}{1.2}
    \setlength{\tabcolsep}{5pt} 
    \caption{Indirect GRF Estimation in Related Works}
    \begin{tabular}{@{} l cccc@{}}
        \toprule
        {Studys} & {Sensors} & {Participants} & {Task} & {Methods} \\
        \midrule
        Refai~\cite{refai2020portable} & 1 IMU & 8 Health & VGW & EEKFs \\
        Eltoukhy~\cite{eltoukhy2017prediction} & Kinect & 9 PD & FGW & Mm \\
        Liu~\cite{Liu2023} & 2 IMUs & \makecell{10 Health\\10 Stroke\\11 PD} & FGW & NEe \\
        Shahabpoor~\cite{shahabpoor2018real} & 2 IMUs & 6 Health & VGW & Rm \\
        Hossain~\cite{hossain2023estimation} & 3 IMUs & \makecell[l]{Dataset A-20 Health \\ Dataset B-17 Health} & VGW & MLa \\
        Bach~\cite{bach2022predicting} & 2 IMUs & 21 Health & VGW & MLa \\
        Kerns~\cite{kerns2023effect} & 3 IMUs & 8 Health & Jumping & Rm \\
        Scheltinga~\cite{scheltinga2023estimating} & 3 IMUs & 12 Health & Running & MLa \\
        Xian~\cite{xian2024imu} & 3 IMUs & 6 Health & VGW & EEKFs \\
        Song~\cite{song2024estimating} & 1-3 IMUs & 44 Health & Running & MLa \\
        \bottomrule
    \end{tabular}
    \caption*{PD: Parkinson's disease; FGW: Fixed-Gait Walking; VGW: Variable-Gait Walking; Rm: Regression model; NEe: Newton–Euler equations; MLa: Machine learning algorithm; EEKFs: Extended Kalman filters; Mm: Musculoskeletal model.}
    \label{tab:study_summary}
\end{table}

The number and placement of wearable IMUs vary significantly across existing studies, substantially influencing the accuracy of GRF estimation. For instance, Song et al.~\cite{song2024estimating} demonstrated that during running tasks in healthy individuals, placing IMUs on both the sacrum and left/right shank got optimal accuracy. Similarly, Sun et al.~\cite{sun2023real} investigated drop-landing tasks and found that the chest position was optimal when only one IMU was employed. Their results further indicated that employing four IMUs (chest, foot, shank, and thigh) achieved performance comparable to configurations using up to eight IMUs. While previous studies have demonstrated the high accuracy and promise of machine learning algorithms and explored various sensor configurations, they were significantly limited to data from healthy individuals. As clinical studies have shown that gait patterns in PD significantly differ from those of healthy individuals\cite{farashi2020distinguishing}, it remains unclear whether these algorithms can achieve comparable accuracy when applied to pathological gait associated with PD. Furthermore, it is uncertain whether the IMU configurations differ for pathological gait, and whether distinct sensor placements significantly affect the accuracy of GRF estimation in Parkinsonian gait.

In summary, several important gaps remain in IMU-based GRF estimation for Parkinsonian gait. First, most existing studies have focused on healthy populations, whereas evidence in pathological gait remains limited. The greater inter-subject heterogeneity and stride-to-stride variability observed in PD make accurate GRF estimation more challenging, and the robustness of existing IMU-based approaches in this population remains insufficiently established. Second, among the two main methodological paradigms, physics-based approaches often rely on predefined biomechanical assumptions and therefore may be challenged when estimating bilateral GRFs in pathological gait across different walking tasks. In contrast, the performance of data-driven deep learning approaches in this context has not yet been systematically evaluated. Third, there is still limited understanding of how IMU number and placement affect deep learning-based GRF estimation in Parkinsonian gait. Although several studies have considered selected sensor setups, they have largely examined fixed configurations rather than providing a systematic evaluation of performance across different sensor counts and placements.

To bridge these gaps, we developed a deep learning framework to estimate bilateral vGRFs from wearable IMU data in Parkinsonian gait. Within the domain of quantified gait analysis, deep learning approaches have been widely explored in healthy populations \cite{khan2024deep}. Among them, convolutional neural network–bidirectional long short-term memory (CNN-BiLSTM) has demonstrated strong potential in capturing both spatial and temporal dependencies in human motion \cite{LIANG2023105372,s25041249}. However, their application to pathological gait in PD remains limited. Therefore, this study aimed to develop and evaluate a CNN-BiLSTM model for bilateral vGRF estimation across walking tasks in individuals with PD. The model was assessed in both PD and healthy control (HC) cohorts, compared with several baseline models within the PD cohort, further examined through cross-dataset external validation, and used to analyze reduced IMU configurations for efficient pathological gait assessment.

\section{Methods}
This study proposed a CNN-BiLSTM based sequence-to-sequence regression framework for synchronous estimation of bilateral vGRF from wearable IMU sensors. The framework was trained and evaluated using gait data from HC and PD participants under both intra-subject and inter-subject settings. In addition to experiments on the primary wearable IMU dataset, cross-dataset external validation was performed on another independent PD gait dataset. Baseline model comparisons and sensor configuration analyses were subsequently conducted within the PD cohort under the inter-subject setting.

\subsection{Dataset}
The primary dataset for model development was a large, multi site, multi task publicly available gait dataset comprising individuals with PD and age matched HC\cite{RF1}. It included wearable IMU data from 13 sensors and pressure data of both feet from the pressure walkway, collected from a total of 126 participants: 61 with PD (25 females, 36 males; age: 76 $\pm$ 9 years) and 65 HC (40 females, 25 males; age: 76 $\pm$ 9 years). In the PD cohort, the modified Hoehn \& Yahr score was 2.15 $\pm$ 0.48, indicating predominantly mild disease severity. And the MDS-UPDRS Part III scores ranged from 6 to 46, with a mean of 24 $\pm$ 10, reflecting mild to moderate motor impairment. Participants performed five walking tasks that fell into three task categories: self-paced walking, hurried-paced walking, and the Timed Up and Go (TUG) test. Self-paced and hurried-paced walking were each conducted both across a gait mat and entirely on it, whereas the TUG test assessed transitional movements from sitting to walking and back. The specific task procedures adhered to the protocols detailed in the referenced study~\cite{RF1}. To enhance generalizability and better reflect real-world application scenarios, data from different tasks were not analyzed separately. Each pass was defined as a complete traversal along the walkway. The anatomical locations of the IMU sensor and their abbreviations are summarized in Table \ref{tableabbr}.

\begin{table}[htbp]
\centering
\caption{The Anatomical Positions of the IMU Sensor Placement and Their Corresponding Abbreviations}
\label{tableabbr}
\begin{tabular}{ll}
\toprule
{Sensor Position} & {Abbreviation} \\
\midrule
Mid-forehead & FH \\
Xiphoid process of the sternum & XP \\
Lower back at vertebral level L4/L5 & LB \\
Right wrist (midway between ulnar and radial styloid) & RW \\
Left wrist (midway between ulnar and radial styloid) & LW \\
Right lateral thigh & RT \\
Left lateral thigh & LT \\
Right lateral shank & RS \\
Left lateral shank & LS \\
Right ankle & RA \\
Left ankle & LA \\
Right dorsum of foot (50\% of foot length) & RD \\
Left dorsum of foot (50\% of foot length) & LD \\
\bottomrule
\end{tabular}
\end{table}

For independent cross-dataset external validation, we further used a separate public PD gait dataset \cite{Shidadata}. This dataset included 26 individuals with PD and provided full body motion capture data and force plate measurements during walking at a self-selected comfortable speed. Kinematic data were recorded at 150 Hz and ground reaction force data at 300 Hz, with 20 walking trials collected per participant. Because wearable IMU recordings were not available in this external dataset, virtual IMU signals were generated from the motion capture data. As head markers were unavailable, only 12 virtual IMUs could be simulated, corresponding to the 12 non-head sensor locations shared with the primary dataset.

\subsection{Data Preprocessing}
For the primary wearable IMU dataset, the IMU data were internally sampled at 1000 Hz, low-pass filtered at 184 Hz, and transmitted at 100 Hz. All sensor modalities were synchronized and recorded at a unified sampling frequency of 100 Hz. In this study, a gait sequence was defined as a continuous walking segment containing alternating foot contacts and typically spanning multiple steps. Sequence onset was identified by the first detectable foot contact, and sequence termination was defined as the subsequent absence of bilateral foot contact, indicating the end of the walking trial. Manual inspection confirmed that no simultaneous bilateral foot-off intervals occurred in this dataset during fast walking. To prevent data leakage, subject-wise and trial-wise data splitting was performed before sliding window segmentation, such that no temporal overlap was shared across the training, validation, and test sets. Continuous gait sequences were then segmented using a sliding window of 64 time points with 25\% overlap. The choice of segment length and overlap ratio was derived from prior studies, which suggested optimal window durations between 0.25 and 1.25 seconds for activity monitoring~\cite{RF2,hossain2023estimation}. Furthermore, both existing literature and preliminary analysis indicated that the stance phase duration in patients with PD typically ranges from 0.5 to 0.8 seconds~\cite{RF4}. These considerations guided the final segmentation strategy. Segments that did not meet the required sequence length were excluded.

For the independent external dataset, only motion capture and force plate recordings were available. Virtual IMU signals were therefore generated from the motion capture data using OpenSim (version 4.5) for subject-specific musculoskeletal scaling and inverse kinematics. Triaxial acceleration and angular velocity signals were then exported at anatomically corresponding sensor locations and processed to be consistent with the signal representation used in the primary dataset. The resulting virtual IMU signals were temporally aligned with the bilateral vertical ground reaction forces obtained from the force plates and resampled to 100 Hz. The same quality control procedures, sliding window segmentation, and body weight normalization applied to the primary dataset were subsequently used for the external dataset.

The model inputs consisted of triaxial acceleration and angular velocity signals from each IMU. Magnetometer signals were excluded due to their susceptibility to environmental magnetic disturbances, which could compromise signal reliability in uncontrolled settings. Recordings without the full 13 IMU sensor set were excluded. After preprocessing, the input data were organized as a tensor of size $(N, T, C_{1})$, where $N$ denotes the number of segments, $T = 64$ is the temporal window length, and $C_{1}$ is the number of input channels. For the primary dataset, $C_{1} = 78$, corresponding to 13 IMUs with six channels per sensor. For the external dataset, $C_{1} = 72$, corresponding to 12 virtual IMUs. For both datasets, the same CNN-BiLSTM backbone architecture was preserved, and only the first convolutional layer was adjusted to match the corresponding input dimensionality. The same strategy was also used in the subsequent ablation experiments with different IMU numbers and configurations, where the selected sensor channels were used as model input, and only the input channel dimension of the first convolutional layer was modified accordingly, while the remaining network architecture was kept unchanged. The model output was defined at the same temporal dimension and represented as $(N, T, C_2)$, where $C_2 = 2$ denotes the left and right vGRF components.

For the primary dataset, vGRFs were derived from pressure and contact area data collected using the pressure walkway. Calibration was based on static standing trial, in which the relationship between contact area and the number of activated pressure sensors was established. This relationship was then applied to the walking trials to estimate contact area dynamically based on sensor activation patterns. Validation results indicated that the vGRF estimated using pressure data and contact area exhibited a coefficient of variation of less than 0.03, demonstrating consistency and controllable error. For the external dataset, vGRFs were obtained from force plate recordings. In both datasets, vGRFs were normalized to each participant's body weight to reduce inter-subject variability. Inspection of the primary dataset showed complete IMU and vGRF data for all PD patients, and complete vGRF data for HC. However, IMU data from HC had a 0.28\% missing rate at the time point level. Short missing intervals of up to five consecutive frames were linearly interpolated, and segments containing longer gaps were excluded.

\subsection{Data Partitioning and Evaluation Protocol}

Two evaluation settings were considered for the primary dataset: intra-subject and inter-subject evaluations. In the intra-subject evaluation, model performance was evaluated using subject-specific partitioning within each cohort (HC or PD). For each participant, walking trials were first split at the trial level into training/validation (80\%) and testing (20\%) subsets before any sliding window segmentation was performed. Each complete trial was assigned exclusively to a single subset, and all windowed segments derived from that trial remained within the same subset. This procedure preserved subject and trial boundaries during model development and prevented temporal leakage across splits. Under this setting, the HC cohort contributed 1,636 training/validation segments and 410 testing segments, whereas the PD cohort contributed 1,424 and 356 segments, respectively. 

In the inter-subject setting, model performance was evaluated on previously unseen individuals. Participants within each cohort were first divided into disjoint training/validation (80\%) and testing (20\%) groups at the subject level, ensuring that no participant appeared in both sets. Gait segmentation into time windows was then performed after this subject-wise split, so that all sequences from a given subject resided entirely within either the training or testing set. This configuration better simulates realistic deployment conditions, where no prior data are available for new users. Under this setup, the HC group provided 2,023 samples for training/validation and 623 for testing, while the PD group contributed 1,460 and 320 samples, respectively.

For cross dataset external validation, the independent external dataset, which included only participants with PD, was evaluated under an inter-subject protocol. Subjects were first partitioned into two disjoint groups: a transfer learning pool and an independent outer test set, with no subject overlap between them. After sliding window segmentation, the transfer learning pool yielded 1,339 training/validation segments, whereas the held-out outer test set contained 425. The transfer learning strategy and domain adaptation strategy were used for model adaptation to reduce the discrepancy between virtual IMU signals and real wearable IMU signals. The pool was divided into training and validation subsets, whereas the outer test set was reserved exclusively for final external evaluation. The CNN-BiLSTM model was initialized using pretrained weights obtained from the primary wearable IMU dataset. Because only 12 virtual IMUs were available in the external dataset, external adaptation and evaluation were restricted to the 12 non-head sensor locations shared by both datasets, yielding an input dimensionality of 72 channels while preserving the remaining network architecture. 

\subsection{Model for Estimation}
\begin{figure}[!t]
\centerline{\includegraphics[width=\columnwidth]{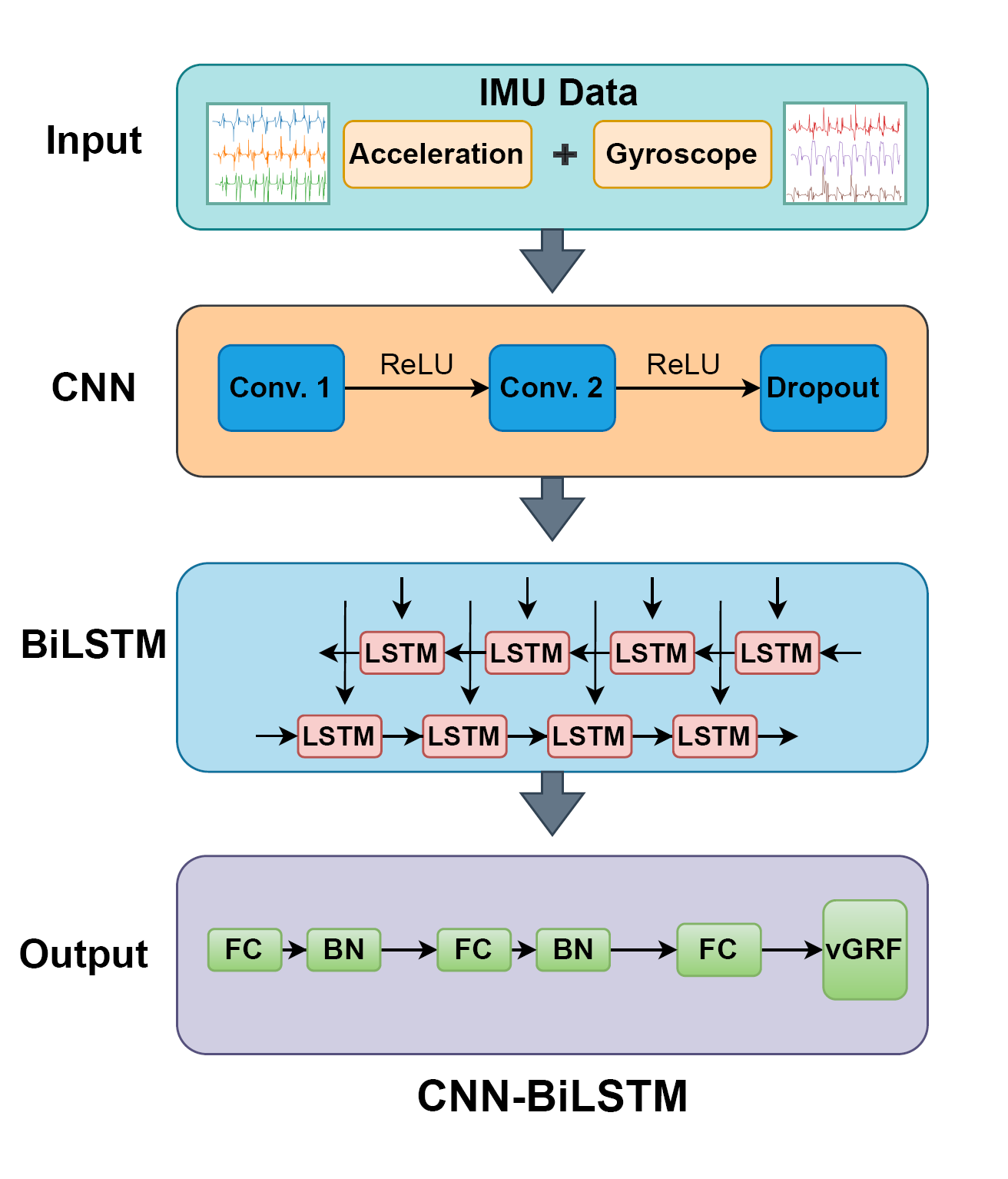}}
\caption{
Architectural representation of the CNN-BiLSTM model. 
CNN denotes Convolutional Neural Network; BiLSTM refers to Bidirectional Long Short-Term Memory; Conv. indicates Convolutional Layer; ReLU stands for Rectified Linear Unit; LSTM is Long Short-Term Memory; FC represents Fully Connected Layer; BN means Batch Normalization.
}
\label{figArchitecture}
\end{figure}

The CNN-BiLSTM is a hybrid neural network architecture specifically designed for sequential estimation tasks involving temporal data with embedded spatial features. It has been widely applied in domains such as activity recognition, joint kinematics estimation, and GRF estimation~\cite{LIANG2023105372,s25041249,RF5,RF6,RF8}. The model proposed in this study, as shown in Fig. \ref{figArchitecture}, comprises three primary modules: (1) a CNN-based feature extraction module, which transforms IMU data into compact, lower dimensional representations; (2) a BiLSTM memory module, which captures temporal dependencies by processing the sequence bidirectionally; and (3) a set of fully connected (FC) layers, responsible for producing the final vGRF estimations. Subsequently, several baseline models were constructed for comparison with the proposed model, including a Multi Layer Perceptron (MLP) \cite{mao2023hybrid}, a Temporal Convolutional Network (TCN) \cite{guo2023speed}, a Transformer model \cite{ramani2024imuoptimize}, the CNN component from the front part of the proposed model, and the BiLSTM component from the latter part.

\subsection{Training, Testing and Parameter Tuning}
Models were trained and optimized using gait data in the primary dataset. Hyperparameter optimization was performed on the training/validation set only using the Optuna framework with 5-fold cross validation\cite{RF10}. To ensure fair comparison across different IMU configurations, the CNN-BiLSTM architecture and the optimized hyperparameter settings were kept fixed across configurations. For each sensor configuration, models were retrained independently, with only the input dimensionality of the first layer adjusted accordingly.

To improve model convergence and generalization, all input features were standardized using z-score standardization. The mean and standard deviation (SD) were computed for each input channel using the training dataset only, and these statistics were reused to transform the validation and test data, thereby preventing information leakage. Model training used the Adaptive Moment Estimation with Decoupled Weight Decay (AdamW) optimizer with a learning rate and weight decay of 0.0001. A batch size of 64 and a maximum of 200 training epochs were employed. Early stopping based on validation loss was used to prevent overfitting, and dropout was applied to the fully connected layers for additional regularization. For all baseline models, hyperparameter optimization was also performed using the Optuna framework prior to training\cite{RF10}. All models employed early stopping and regularization techniques to prevent overfitting. A summary of the optimized hyperparameters of the CNN-BiLSTM model is provided in Table ~\ref{tablehyperparameters}.

\begin{table}[h!]
    \captionsetup{
    justification=centering, 
    singlelinecheck=false, 
    labelfont=bf, 
    labelsep=space 
    }
    \caption{Selected Hyperparameters for the CNN-BiLSTM Model.}
    \centering
    \begin{tabular}{ll} 
    \toprule
    \multicolumn{2}{c}{\textbf{Hyperparameters}} \\ \midrule  
    CNN1D-1 kernel size (3,1)  & CNN1D-1 Output Channels 256  \\
    CNN1D-2 kernel size (3,1)  & CNN1D-2 Output Channels 128  \\
    LSTM Hidden Size 256       & LSTM Num Layers 3 \\
    FC-1 Output Size 60         & FC-2 Output Size 30 \\
    Dropout Rate 0.1           & \\ \bottomrule 
    \end{tabular}
    \caption*{CNN1D refers to a one-dimensional convolutional layer; LSTM stands for long short-term memory; FC denotes a fully connected layer.}
    \label{tablehyperparameters}
\end{table}

For cross dataset external validation, transfer learning was performed by initializing the CNN-BiLSTM model with pretrained weights obtained from the primary dataset and subsequently adapting it on the external training subset. The same training and validation strategy was applied to the external dataset. Standardization parameters were estimated from the external training subset only and then applied to the corresponding validation and held-out test subsets. The final adapted model was evaluated on the independent external test subjects.

\subsection{Statistical Tests and Ablation Study}
Model performance was assessed using mean absolute error normalized to body weight [MAE (\%BW)], relative root mean square error (rRMSE), and the coefficient of determination ($R^2$) between measured and estimated vGRF. For each segmented sample, MAE and $R^2$ were first computed separately for the left and right vGRF channels and then averaged to obtain segment-level values. These segment-level metrics were subsequently aggregated within each subject and then summarized at the cohort level to obtain group statistics. Results are reported as mean $\pm$ SD or median [interquartile range (IQR)], as appropriate to the data distribution. Statistical analyses were conducted by first assessing data normality (Shapiro–Wilk test) and variance homogeneity (Levene's test). Based on these results, one way ANOVA with Tukey's post-hoc test (normal and homogeneous data), Welch's ANOVA with Games-Howell post-hoc test (normal but heterogeneous data), or Kruskal–Wallis with Dunn's test (non-normal data) was applied. Given the observed variance heterogeneity, Games-Howell tests were primarily used. Statistical significance was set at $p \leq 0.05$.

Because exhaustive ablation across all 13 IMUs would yield 8,191 possible sensor combinations, an initial screening step was performed to reduce the computational burden. Specifically, each of the 13 single-IMU configurations was first evaluated to assess whether a single sensor could provide sufficient information in the PD cohort under the inter-subject setting. In accordance with prior algorithm development studies\cite{RF11}, which consider an $R^2$ exceeding 0.80 indicative of high estimation accuracy, IMU placements failing to meet this criterion were excluded. As a result, 10 IMUs, with a total of 1,023 IMU combinations, were selected for the subsequent ablation study. 

\section{Results}
\subsection{Intra- and Inter-Subject Variability in vGRF Estimation for HC and PD}

This study first evaluates the performance of the model in estimating vGRF using the primary dataset from all 13 IMUs. The analysis considers two cohorts, HC and patients with PD, under both intra- and inter-subject conditions. The results are summarized in Table~\ref{tableThirteenIMUs}.

In the intra-subject analysis, where the model was trained and tested on data from individuals within the same group to evaluate population-specific estimation performance, the following results were obtained. For HC, the model achieved an MAE of $0.91 \pm 0.87\%$ BW, and an \( R^2 \) of \( 0.98 \pm 0.08 \). For PD patients, the model produced an MAE of $1.14 \pm 0.95\%$ BW, and an \( R^2 \) of \( 0.98 \pm 0.06 \). The difference between the two groups was not statistically significant (\( p > 0.05 \)).

In the inter-subject analysis, where the model was trained on data from multiple individuals and tested on previously unseen subjects to evaluate generalizability, a decrease in estimation performance was observed. For HC, the model achieved an MAE of $2.71 \pm 2.47\%$ BW, and an \( R^2 \) of \( 0.93 \pm 0.25 \). For PD patients, the model achieved an MAE of $3.99 \pm 2.63\%$ BW, and an \( R^2 \) of \( 0.91 \pm 0.10 \). These results indicate a higher estimation error and reduced model accuracy in the PD group compared to the HC.

When comparing intra- and inter-subject conditions, the latter demonstrated significantly lower estimation accuracy, as indicated by increased MAE and reduced \( R^2 \) (\( p < 0.05 \)). Despite this decline, the \( R^2 \) remained above 0.90 in both groups, indicating that the model retained a strong capacity to capture essential vGRF patterns across individuals.

\begin{table}[h!]
\captionsetup{
    justification=raggedright, 
    singlelinecheck=false, 
    labelfont=bf, 
    labelsep=space 
}
\caption{Estimation Performance of vGRF Using Thirteen IMUs: Intra- and Inter-subject Comparisons Between Healthy Controls and PD Patients}
\centerline{\includegraphics[width=\columnwidth]{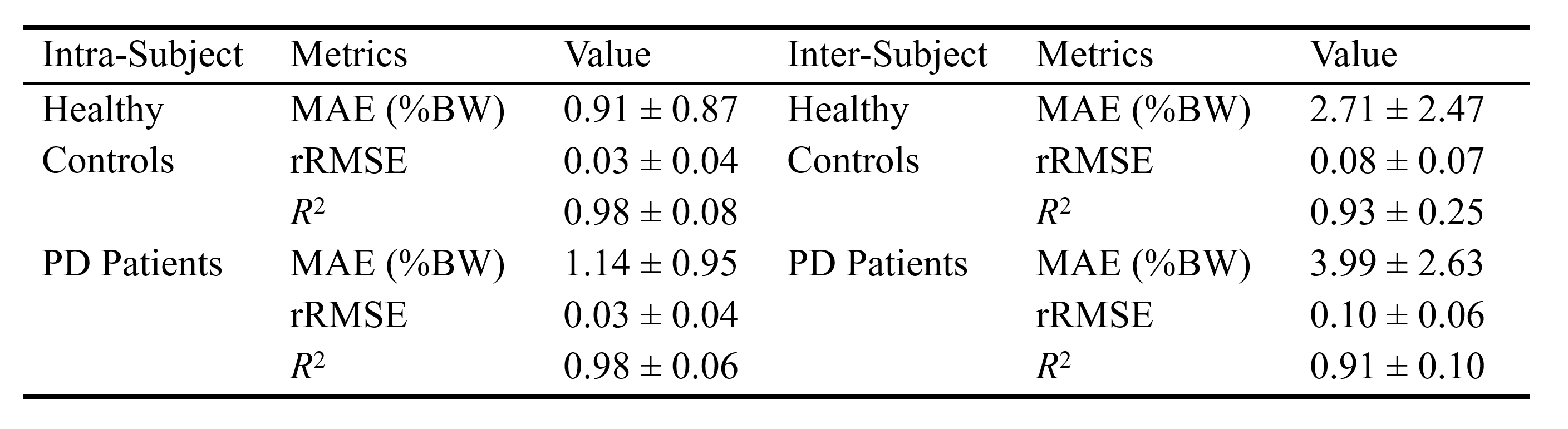}}
\captionsetup{font=small} 
\captionsetup{justification=raggedright}
\caption*{MAE: Mean Absolute Error; rRMSE: Relative RMSE; $R^2$: Coefficient of Determination. MAE values are expressed as percentages of body weight (\%BW).}
\label{tableThirteenIMUs}
\end{table}

\subsection{Baseline Model Comparison and External Validation for Inter-Subject PD Evaluation}

To evaluate the performance of the proposed CNN-BiLSTM model, we first compared it with five baseline models on the primary PD dataset under the full 13-IMU inter-subject setting, including TCN, MLP, Transformer, and two ablated variants derived from the proposed architecture, namely CNN-only and BiLSTM-only. All models were evaluated under consistent experimental settings. As summarized in Table~\ref{tablebaseline}, the CNN-BiLSTM model consistently outperformed all other models across the three evaluation metrics. Although the Transformer model demonstrated relatively competitive results (MAE: $3.80 \pm 2.41\%$ BW; $R^2$: $0.90 \pm 0.11$), its performance variability suggests reduced robustness. The ablated variants, CNN-only and BiLSTM-only, derived from the proposed framework, exhibited diminished performance when applied independently, with $R^2$ scores of $0.89 \pm 0.09$ and $0.88 \pm 0.13$, respectively. In contrast, traditional baselines such as MLP and TCN yielded the lowest performance. These results support the advantage of combining convolutional feature extraction with bidirectional temporal modeling for inter-subject vGRF estimation in patients with PD.

\begin{table}[h!]
\captionsetup{
    justification=raggedright, 
    singlelinecheck=false, 
    labelfont=bf, 
    labelsep=space 
}
\caption{Performance Comparison Between the Proposed CNN-BiLSTM Model and Baseline Models}
\centerline{\includegraphics[width=\columnwidth]{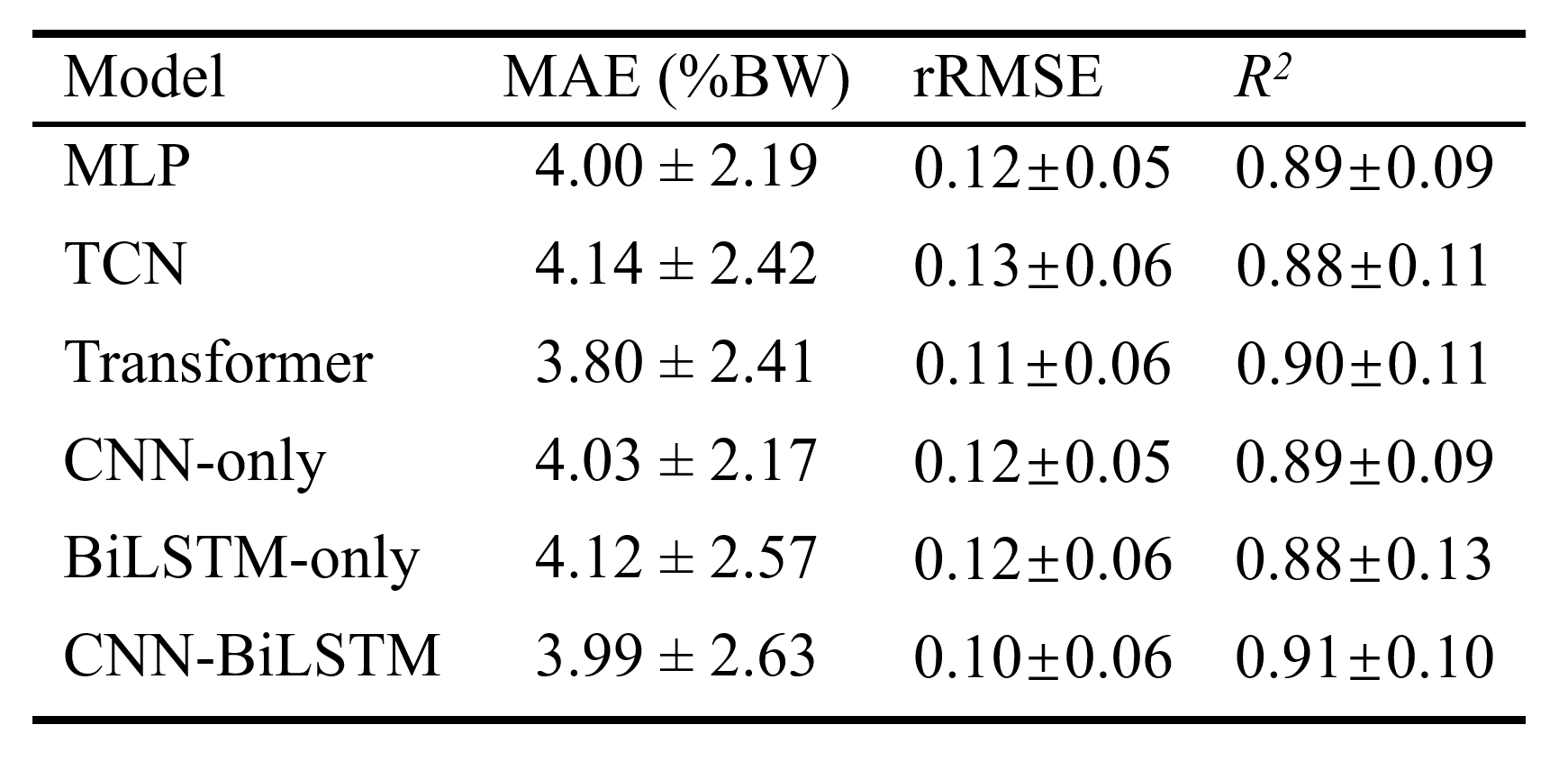}}
\captionsetup{font=small} 
\captionsetup{justification=raggedright}
\caption*{
MLP: Multi-Layer Perceptron; TCN: Temporal Convolutional Network; CNN-only: convolutional component of the proposed CNN-BiLSTM model; BiLSTM-only: bidirectional long short-term memory component of the proposed CNN-BiLSTM model; MAE: Mean Absolute Error; rRMSE: Relative RMSE; $R^2$: Coefficient of Determination. MAE are expressed as percentages of body weight (\%BW).}

\label{tablebaseline}
\end{table}

For cross dataset external validation, a direct comparison with the 13-IMU primary setting was not feasible because IMU on the head was unavailable in the external dataset. Therefore, both datasets were restricted to the 12 shared non-head IMUs to enable a matched comparison. The results of the comparison between the primary and external datasets are summarized in Table~\ref{tableexternal}. Under this 12-IMU setting, the model achieved an MAE of $4.08 \pm 2.49\%$ BW, and a $R^2$ of $0.88 \pm 0.11$ on the primary dataset, compared with an MAE of $7.45 \pm 3.85\%$ BW, and a $R^2$ of $0.86 \pm 0.17$ on the external dataset. No statistically significant difference was detected in $R^2$ between the matched 12-IMU primary dataset results and the external validation results ($p = 0.51$), supporting the cross dataset generalizability of the model.

\begin{table}[h!]
\captionsetup{
    justification=raggedright, 
    singlelinecheck=false, 
    labelfont=bf, 
    labelsep=space 
}
\caption{Comparison of vGRF Estimation Performance Between the Primary and External Datasets Using 12 Matched IMUs}
\centerline{\includegraphics[width=\columnwidth]{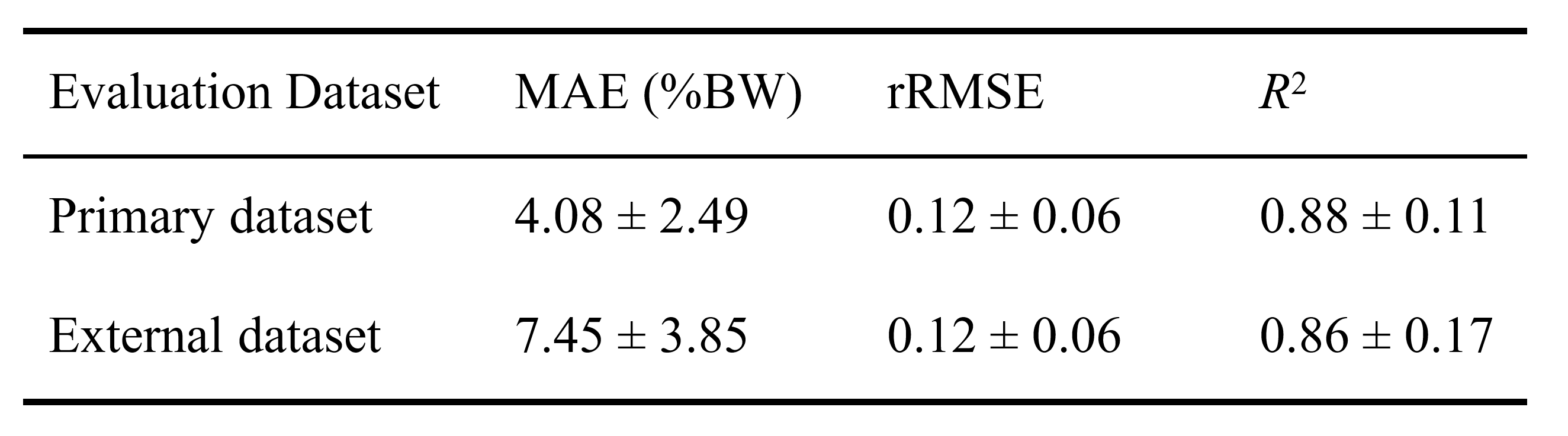}}
\captionsetup{font=small} 
\captionsetup{justification=raggedright}
\caption*{MAE: Mean Absolute Error; rRMSE: Relative RMSE; $R^2$: Coefficient of Determination. MAE values are expressed as percentages of body weight (\%BW).}
\label{tableexternal}
\end{table}

\subsection{Impact of Single IMU Placement on vGRF Estimation in HC and PD Patients}
\begin{figure}[!t]
\centerline{\includegraphics[width=\columnwidth]{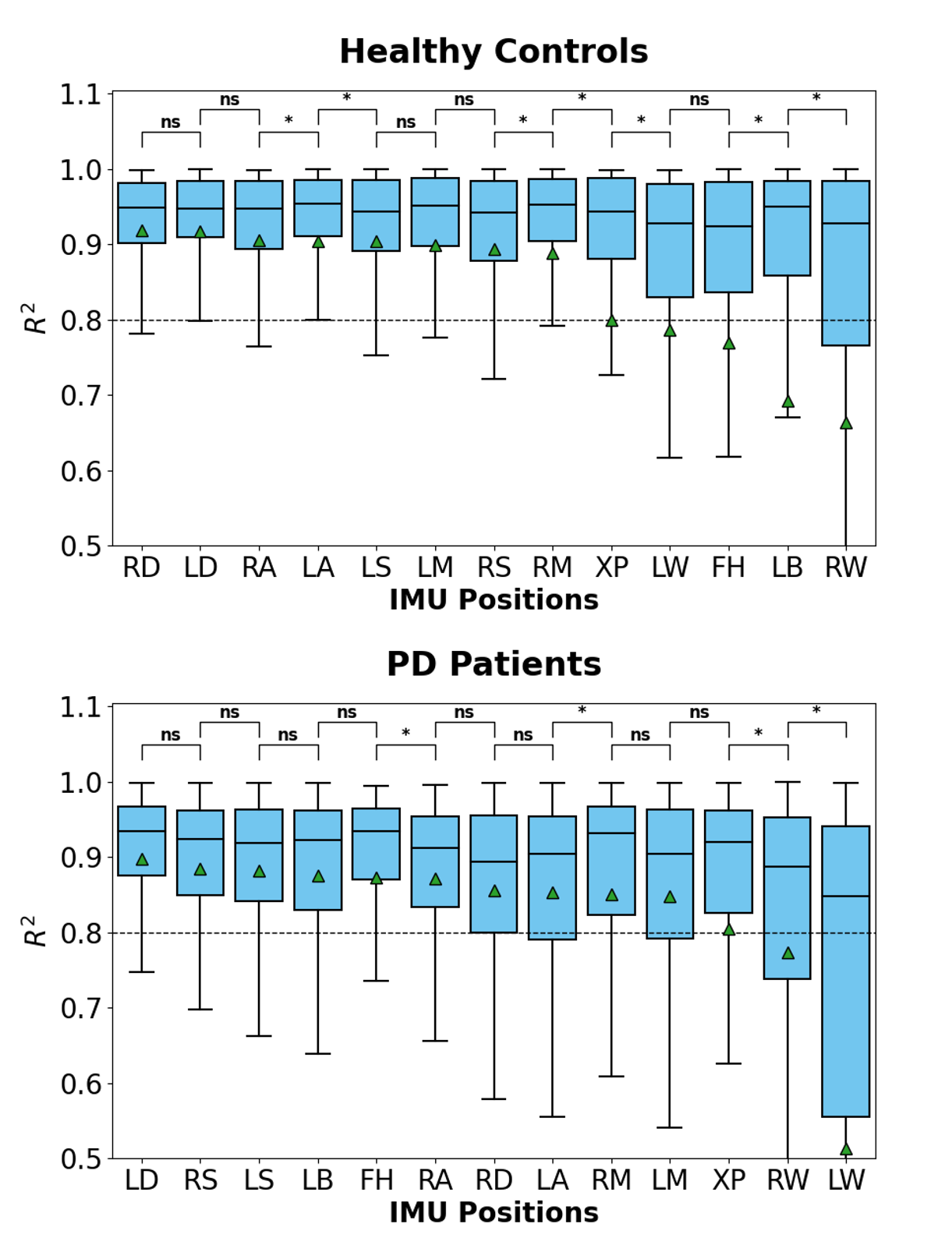}}
\caption{The vGRF estimation accuracy for healthy controls and PD patients using a single IMU placed at 13 different anatomical positions: Forehead (FH), Xiphoid Process (XP), Left Ankle (LA), Right Ankle (RA), Left Dorsal Foot (LD), Right Dorsal Foot (RD), Left Lateral Shank (LS), Right Lateral Shank (RS), Left Mid-Lateral Thigh (LM), Right Mid-Lateral Thigh (RM), Left Wrist (LW), Right Wrist (RW), and Lower Back (LB). In the box plot, the lines within each box indicate the median accuracy, while the triangular markers represent the mean accuracy. Statistical significance ($*$ for $p \leq 0.05$; \textit{ns} for no significant) was assessed via the Games-Howell post-hoc test after normality (Shapiro–Wilk) and variance homogeneity (Levene) testing. Comparisons were made between adjacent IMU rankings based on mean and median accuracy.}
\label{figHCVSPD}
\end{figure}

The placement of the IMU plays a critical role in feature extraction, directly influencing the accuracy of model estimation~\cite{RF12}. To assess this effect, we evaluated the CNN-BiLSTM model's performance in estimating vGRF using a single IMU positioned at thirteen different anatomical positions across inter-subject, including both HC and patients with PD. The results, ranked from highest to lowest accuracy, are illustrated in Fig.~\ref{figHCVSPD}.

Overall, the $R^2$ were higher and standard deviations were lower in the HC compared to patients with PD, indicating greater estimation consistency and accuracy in the HC. Among HC, the two highest performing IMU placements, RD and LD, showed no statistically significant difference (\(p>0.05\)). Similarly, in the PD patients, the five highest performing placements, LD, RS, LS, LB, and FH, also exhibited no statistically significant differences (\(p>0.05\)). Statistical comparisons between consecutively ranked sensor placements revealed two distinct accuracy drop-off points in both groups. In HC, we observed minimal differences between the left and right sides of the same body part, whereas in PD patients, the side-to-side differences were more pronounced. Meanwhile, the ranking of estimation accuracy from highest to lowest differed significantly between the two groups. In HC, the RD and LD achieved the highest two median accuracies, whereas in PD patients, the LD yielded the best estimation performance. Furthermore, although overall performance was higher in HC, localized deviations were observed at specific placements (e.g., FH and LB), where the mean 
$R^2$ in PD was comparable to or slightly exceeded that in HC. These findings consistently underscore the critical importance of IMU placement. The observed asymmetry between left and right sides may reflect the heterogeneous distribution of motor symptom severity across limbs in patients with PD. Consistent with this interpretation, lower limb laterality, quantified using MDS-UPDRS Part III lower extremity items \cite{seuthe2024gait}, showed a higher prevalence of left dominant impairment (43.1\%) than right dominant impairment (32.8\%) in this cohort, which may partly account for the superior LD performance at the cohort level. These findings also highlight the importance of developing IMU placement strategies specifically tailored for PD.

Notably, in the PD cohort, IMU placements at the XP ($R^2 = 0.76 \pm 0.58$), LW ($R^2 = 0.43 \pm 1.01$), and RW ($R^2 = 0.78 \pm 0.42$) exhibited relatively low estimation accuracy, with $R^2$ falling below 0.8. Considering the increased computational complexity associated with a larger number of IMU inputs, these three positions were excluded from subsequent ablation experiments. As a result, the remaining 10 IMU placements were selected for further analysis.

\subsection{Distribution of vGRF Estimation Performance Across IMU Counts in PD cohort}
\begin{figure}[!t]
\centerline{\includegraphics[width=\columnwidth]{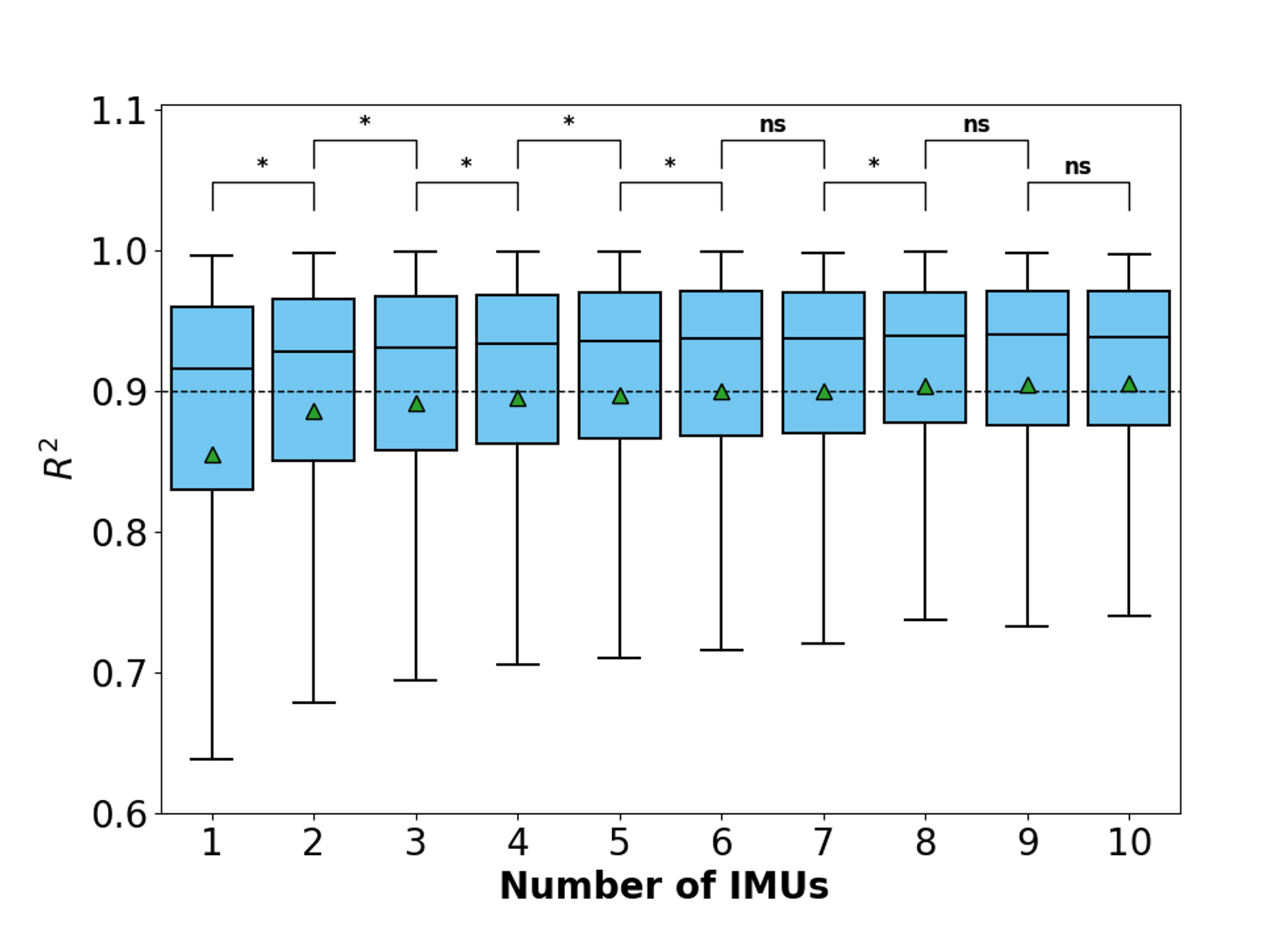}}
\caption{The vGRF estimation distribution, represented by $R^2$, was assessed in patients with PD using models incorporating varying numbers of IMUs, with each group including all possible sensor configurations for a given number of IMUs. In the box plot, lines within the boxes indicate the median accuracy, while triangular markers denote the mean accuracy. Statistical significance ($*$ for $p \leq 0.05$; \textit{ns} for no significant) was assessed via the Games-Howell post-hoc test after normality (Shapiro-Wilk) and variance homogeneity (Levene) testing. For statistical analysis, IMU configurations were ranked incrementally based on mean and median accuracy, with comparisons performed only between adjacent conditions.}
\label{figQuantity}
\end{figure}

To examine how estimation performance varied across sensor placements, we further analyzed the $R^2$ across varying quantities of IMUs by systematically exploring all 1,023 possible combinations of 1 to 10 IMU sensors. For each configuration, the CNN-BiLSTM model was retrained under the same experimental protocol. As shown in Fig.~\ref{figQuantity}, for each IMU count, all possible combinations were evaluated separately, and the reported $R^2$ values represent the distribution of performance across combinations corresponding to that sensor number. The results demonstrate that the proposed model consistently maintained high estimation accuracy, with median $R^2$ values exceeding 0.9 across all tested configurations.

Across all combinations, the performance distributions generally shifted toward higher $R^2$ values as the number of IMUs increased from 1 to 6, after which the improvement became less pronounced. The distributions also became narrower at higher IMU counts, suggesting reduced sensitivity to specific sensor placement when more sensors were available. However, these patterns should be interpreted descriptively, because the all-combination analysis reflects not only the amount of sensing information but also the structure of the configuration space at each count, including the greater likelihood that higher-count configurations contain anatomically informative sensor locations. Notably, substantial variation in performance remained among configurations with the same number of IMUs, underscoring the importance of sensor placement. The largest upward shift in performance was observed when increasing the number of IMUs from 1 to 3, whereas further increases yielded more modest gains.

\subsection{Impact of IMU Quantity on Optimal Placement and Estimation Accuracy in PD Patients}
\begin{figure*}[!t]
\centerline{\includegraphics[width=\textwidth]{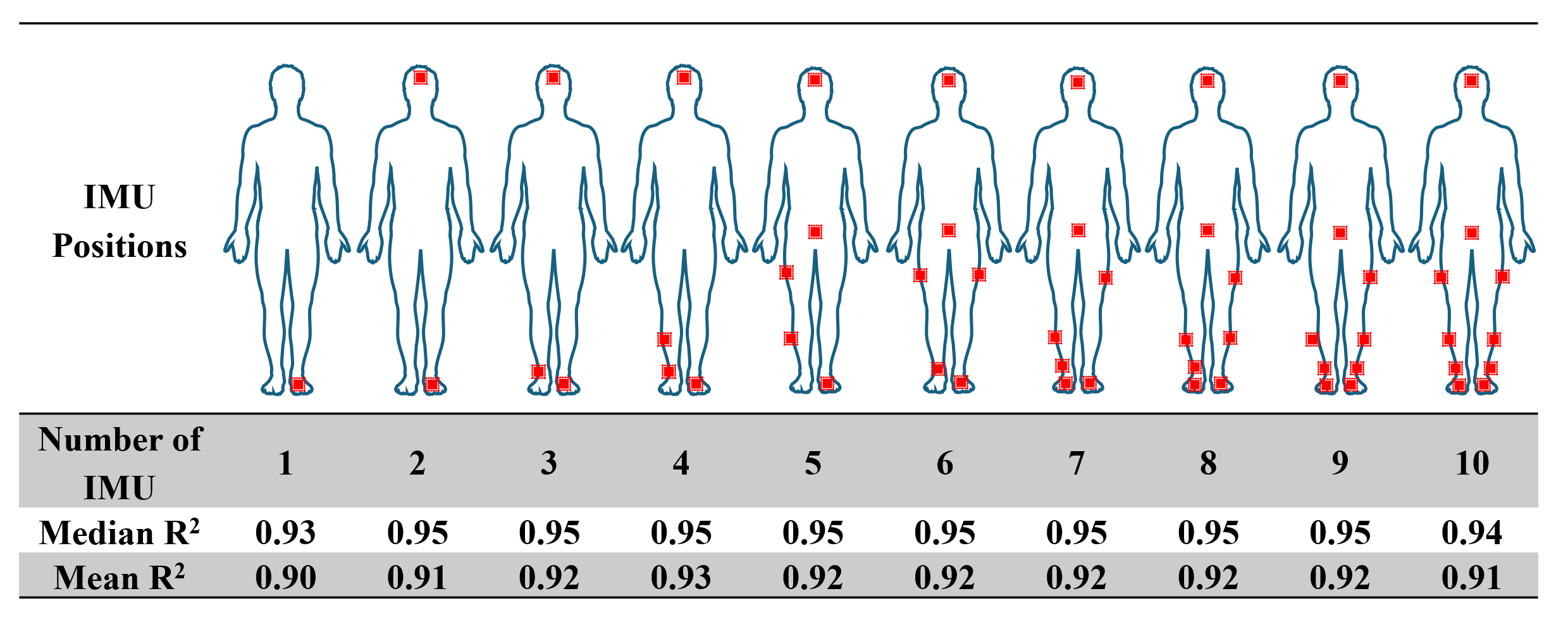}}
\caption{Optimized IMU configurations for different numbers of IMUs in patients with PD.}
\label{figOptimized}
\end{figure*}

\begin{figure*}[!t]
\centerline{\includegraphics[width=\textwidth]{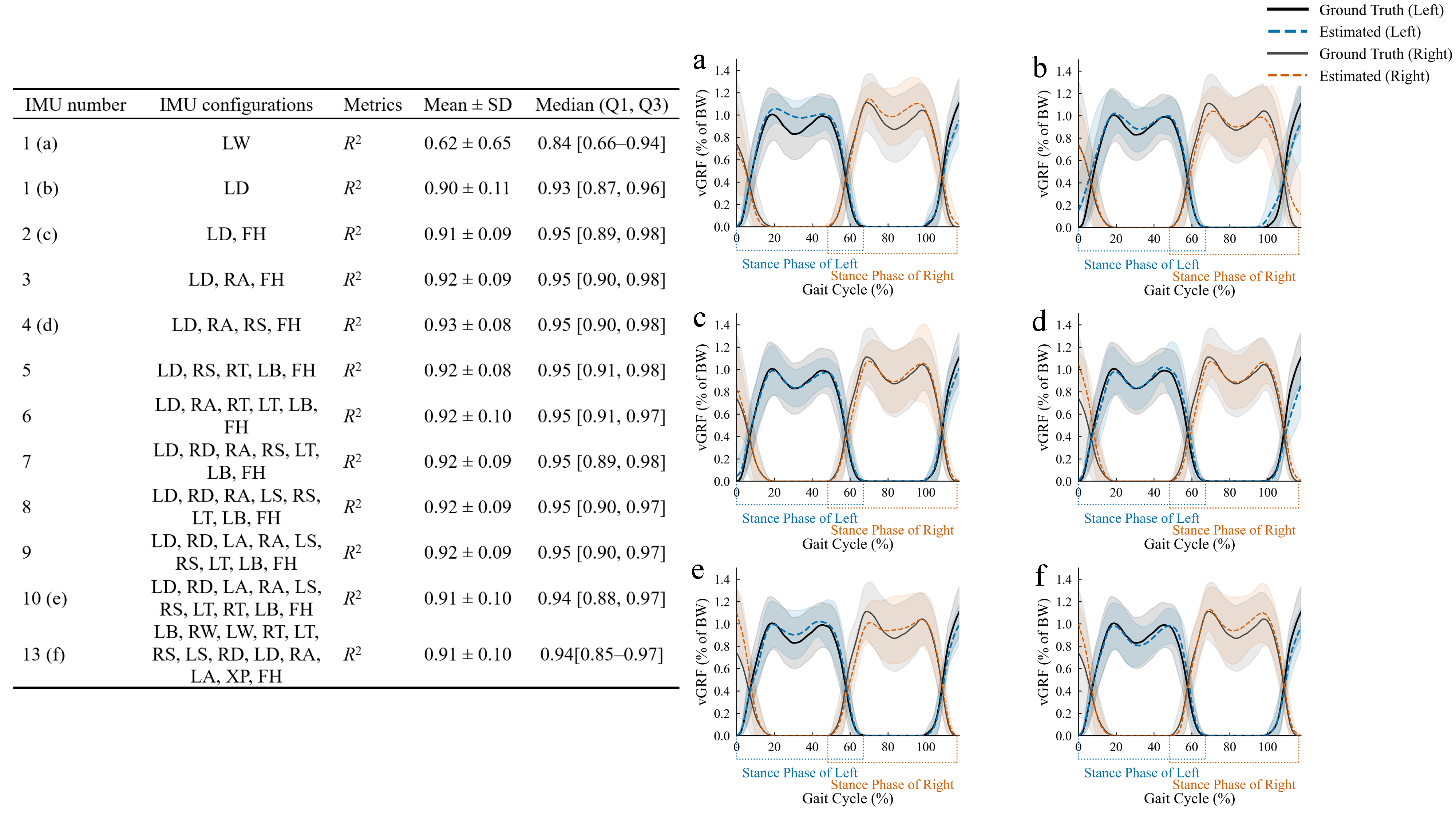}}
\caption{Representative IMU configurations for different sensor counts in patients with Parkinson's disease (PD). The table summarizes the best-performing configuration at each sensor count, alone with the worst-performing single IMU configuration. The results are reported as mean $\pm$ standard deviation (SD) and median (Q1, Q3). Panels (a)–(f) show representative bilateral gait cycle vGRF waveforms (\%BW) for the selected configurations indicated in the table. Panel (a) presents the worst-case example, whereas panel (d) presents the best-case example. Solid lines represent ground truth, dashed lines represent model estimation, and shaded regions denote SD. IMU placements are abbreviated as follows: FH, forehead; XP, xiphoid; LA, left ankle; RA, right ankle; LD, left dorsal foot; RD, right dorsal foot; LS, left lateral shank; RS, right lateral shank; LT, left thigh; RT, right thigh; LW, left wrist; RW, right wrist; and LB, lower back.}
\label{fig6}
\end{figure*}

To determine the optimal IMU placement for varying sensor configurations under realistic generalization conditions, an ablation study was conducted within the PD inter-subject evaluation setting, evaluating a total of 1,023 unique IMU combinations. The optimal configuration for each sensor count was defined as the combination that achieved the highest $R^2$ among all possible configurations with the same number of IMUs. The results are presented in Fig.\ref{figOptimized}, and the corresponding $R^2$ statistics are summarized in Fig. \ref{fig6}, together with representative gait cycle waveforms for the selected IMU configurations.

For single IMU configurations, the optimal placement for maximizing accuracy was at the LD position. However, this configuration also resulted in the lowest median and mean $R^2$ compared to the optimal configurations with a higher number of IMUs, indicating limited estimation reliability in this case. Based on both median and mean accuracy metrics, the best overall configuration was achieved using four IMUs. Adding more IMUs beyond this number did not lead to substantial improvements in estimation performance, as the accuracy tended to plateau. Statistical comparisons of $p$ between adjacent IMU configurations showed no significant differences for setups using 2 to 9 IMUs. However, a significant decrease in accuracy was observed when using only one IMU or increasing the number to ten ($p \leq 0.05$). In summary, for minimal sensor use, a two IMU configuration at FH and LD was recommended. To achieve the highest $R^2$, the optimal configuration consists of four IMUs placed at LD, RA, RS, and FH.

\section{Discussion}

The present study demonstrates that bilateral vGRF estimation in Parkinsonian gait is feasible using wearable IMUs under an inter-subject setting. Despite the inherent variability of pathological gait, the proposed model maintained robust performance. Sensor configuration analysis further suggested that gait characteristics in PD differ from those of healthy controls, while a two-IMU setup could still provide comparable estimation performance. Collectively, these findings underscore the importance of disease-specific validation for the development of wearable gait assessment systems.

\subsection{Principal Outcomes: Generalization and Sensor Efficiency}
In line with the hypothesis, the results demonstrated that deep learning can effectively capture the complex gait dynamics associated with PD, achieving high estimation accuracy ($R^2 > 0.90$) across both HC and PD patients, as well as under varying grouping strategies. The intra-subject analysis revealed comparable performance between PD patients and HC ($R^2 = 0.98$ for both), indicating that the CNN-BiLSTM model successfully captured the inherent variability of pathological gait and mitigated the group-dependent discrepancies often encountered in biomechanical estimation tasks involving neurological disorders. This suggests that, when the training data include patient-specific information, the model can learn individualized gait signatures, providing a solid foundation for developing personalized clinical assessment tools that can be initiated during early stage hospital visits at the onset of care. Conversely, the inter-subject evaluation, while still achieving strong accuracy ($R^2 = 0.91$ for PD and $R^2 = 0.93$ for HC), exhibited a slight reduction in performance, particularly in the PD group. This finding aligns with prior research, which has consistently highlighted the greater heterogeneity in pathological gait compared to healthy patterns. This heterogeneity normally poses significant challenges to model generalization, especially for physics-based approaches \cite{zanardi2021gait}. The reduced inter-subject accuracy in PD patients may be attributed to fluctuating motor symptoms, asymmetrical limb movement, and episodic gait disturbances, such as freezing and staggering factors that are inherently difficult to generalize in the absence of individual calibration\cite{smith2021current}. Nonetheless, the inter-subject approach supports broader deployment scenarios, such as large-scale screening and home-based monitoring, where personalization is not always feasible but datasets from other patient populations are available.

In the baseline model comparison, the proposed CNN-BiLSTM architecture demonstrated clear advantages over traditional models in capturing the complex spatiotemporal characteristics of pathological gait. Although CNN-BiLSTM is not a novel architecture in itself, the systematic comparisons and ablation analyses presented here provide task-specific empirical support for its suitability in bilateral vGRF estimation from wearable IMU data in PD. In particular, ablation results showed that removing either the convolutional or recurrent component was associated with reduced estimation performance, indicating that both components contributed to the final model. The detailed results further suggest that the convolutional module was particularly useful for extracting informative local motion features from IMU signals. In contrast, conventional models such as MLP and TCN showed limited capability in modeling gait dynamics, underscoring the benefits of combining deep spatial and sequential modeling for inter-subject gait estimation in PD patients.

Importantly, the study provides compelling evidence that sensor configuration plays a critical role in estimation performance, particularly under pathological gait conditions. The results revealed significant discrepancies in optimal IMU placements between HC and PD patients. When using only a single IMU, overall accuracy was consistently higher in HC than in PD patients, and the ranked order of sensor effectiveness differed substantially between the two cohorts. Although minor exceptions were observed at specific positions such as FH and LB, these comparisons should be interpreted cautiously because the HC and PD results were ranked independently, and bilateral vGRF estimation from a single IMU remains inherently under-constrained. One possible interpretation is that the greater asymmetry and variability of Parkinsonian gait alter the biomechanical information captured at different sensor locations. A further observation in the PD cohort was that the most informative sensor locations were not bilaterally symmetric. For instance, while the RD and LD achieved the highest two estimation accuracies in HC, only the LD emerged as the most effective placement for PD patients. This pattern may reflect the asymmetric manifestation of pathological gait in PD, although the present study was not designed to determine whether this effect was driven by symptom dominance or other subject-specific factors. Building on these insights, a comprehensive ablation study of 1,023 IMU combinations was conducted to assess the impact of sensor count and placement on the estimation accuracy. The results indicated that estimation accuracy improved steadily up to a four sensor configuration, beyond which additional sensors offered only marginal gains. Notably, a simplified two IMU configuration, positioned at the LD and FH, achieved near optimal performance ($R^2 = 0.91 \pm 0.09$), representing a practical trade-off between estimation accuracy and sensor system streamlining and comfort. These findings offer valuable guidelines for the future design of wearable systems targeting real-world gait analysis in PD, especially in scenarios where reducing sensor burden is critical for ensuring patient compliance and long term usability.

\subsection{Comparison with Prior Studies and Methodologies}
Although numerous studies have investigated GRF estimation using wearable IMUs, relatively few have focused specifically on pathological gait patterns such as those observed in PD \cite{hossain2023estimation,bach2022predicting,shahabpoor2018real,kerns2023effect,scheltinga2023estimating,xian2024imu,song2024estimating}. Existing approaches can be broadly divided into physics-based and data-driven methods. Physics-based methods, such as musculoskeletal modeling or Newton-Euler equations, offer strong biomechanical interpretability but are constrained by multiple assumptions that limit their applicability to pathological gait, especially when estimating bilateral vGRFs during walking \cite{Karatsidis}. To our knowledge, Liu et al.\cite{Liu2023} reported the only prior Newton-Euler-based study that included both healthy participants and patients with PD for vGRF estimation. However, that study was based on a single dataset with 10 healthy participants and 11 patients with PD, whereas the present study used a larger cohort and also included independent external validation, providing a broader basis for assessing robustness across datasets. Under a comparable two-shank IMU configuration, Liu et al. reported an MAE of 5.82-6.36\% BW, whereas our model achieved an MAE of 3.38\% BW. Furthermore, their approach relied on a fixed walking speed, static calibration, and an assumption of gait symmetry, which may be less suitable for the asymmetric and more variable gait patterns characteristic of PD. Consistent with this point, our results showed that sensor performance in the PD cohort was not bilaterally symmetric; for example, the LD configuration showed stronger predictive performance than several alternative single-sensor placements. Although this observation does not establish a direct link to symptom laterality, it is consistent with the broader view that asymmetric biomechanical alterations in PD can influence IMU-based vGRF estimation.

In contrast to physics-based methods, data-driven methods, particularly deep learning, have demonstrated stronger adaptability to complex and variable movement patterns. However, dedicated deep learning studies focusing specifically on pathological gait in PD remain limited. For instance, Sharma et al.\cite{Sharma} used a single torso mounted IMU and applied an LSTM model to estimate vGRF during outdoor walking and running in two healthy participants. Similar to our study, they evaluated both intra-subject and inter-subject settings. In their intra-subject experiments, the model achieved normalized root mean square errors (nRMSE) of 8.38–8.54\%, whereas performance deteriorated by approximately three- to four-fold under inter-subject evaluation, indicating substantial sensitivity to distributional shift. By comparison, our model was trained on a larger and more heterogeneous cohort of 65 elderly healthy controls and maintained good performance in both intra-subject and inter-subject settings, with rRMSE values of 3\% and 8\%, respectively. While nRMSE and rRMSE are normalized using different reference values and are not directly comparable, both metrics reflect the improved robustness of our approach across heterogeneous test conditions. 

Studies examining IMU configuration have also reported findings that are qualitatively consistent with ours. Havashinezhadian et al.\cite{Havashinezhadian} used a reservoir computing neural network to estimate vGRF in 27 healthy individuals and 18 participants with medial knee osteoarthritis using five IMUs placed on the shoe, heel, above the medial malleolus, mid-tibia, and medial knee. They found that the shoe sensor position achieved the best estimation performance, which aligned with our finding that the dorsum sensor configuration achieved the most accurate results. Zhu et al.\cite{Zhutrans} proposed a transformer-based model for vGRF estimation during walking in 10 healthy participants using five IMUs placed on the pelvis, bilateral thighs, and ankles. Their model achieved a mean absolute percentage error (MAPE) of approximately 6.8\%, which increased only modestly to 7.3\% when the configuration was reduced to three IMUs, but rose more noticeably to 8.5\% when only two ankle IMUs were retained. Although these studies differed from ours in cohort characteristics, tasks, and evaluation metrics, their ablation results are qualitatively consistent with our findings, suggesting that a reduced but strategically placed sensor set may preserve useful estimation performance while improving wearability.

Taken together, the present study provides a systematic evaluation of IMU-based bilateral vGRF estimation in Parkinsonian gait using a deep learning framework. The proposed CNN-BiLSTM model reduces reliance on some of the simplifying assumptions commonly required by physics-based methods and was evaluated under both intra-subject and inter-subject settings, with additional cross-dataset external validation. Comparative analyses of single IMU configurations further showed that the sensor locations associated with the best estimation performance differed between the HC and PD cohorts, highlighting the need to consider disease-specific factors when designing wearable gait assessment systems. In addition, the ablation analysis provided quantitative evidence on how sensor number and anatomical placement influence estimation performance in PD. The highest performance was obtained with four IMUs (LD, RS, RT, FH; $R^2 = 0.93$), whereas a reduced two-IMU configuration (LD and FH) maintained similar performance ($R^2 = 0.91$). This lower-burden configuration therefore represents a practical trade-off between estimation performance and wearable simplicity, which may support more feasible real-world deployment.

\subsection{Limitations and Future Work}
This study has several limitations. Although independent cross-dataset external validation was performed, the availability of public PD gait datasets remains limited. In addition, the external dataset relied on motion-capture-derived virtual IMU signals rather than real wearable IMU recordings, which may not fully represent the variability encountered in practical deployment. Further validation on larger, more diverse, and fully wearable datasets is therefore still needed to establish broader generalizability. Second, this study does not incorporate clinical indicators of disease severity. Gait characteristics in PD can vary significantly depending on the severity of motor symptoms \cite{creaby2018gait}. However, clinical severity scores, limb symmetry metrics, and other relevant features are not included in the current analysis. The potential impact of these indicators on model performance remains uncertain and warrants further investigation. Third, the ablation analysis was conducted only under the inter-subject setting and the intra-subject evaluation was based on tasks collected within a single recording session for each participant. As a result, inter-day variability, sensor repositioning effects, and longitudinal changes associated with disease progression were not evaluated. These factors may be particularly relevant for wearable monitoring in progressive conditions such as PD. Finally, although multiple baseline architectures were compared under the full IMU setting, the large-scale sensor-configuration analysis was conducted only using the CNN-BiLSTM model. Consequently, the potential interaction between model architecture and optimal sensor configuration was not systematically examined and remains an important direction for future investigation.

Future work should address these limitations by extending the available datasets to include more diverse walking tasks, broader participant populations, and repeated recordings across sessions. Incorporating clinical descriptors such as disease severity scores may enhance model accuracy and clinical relevance. Furthermore, exploring transfer learning techniques may improve model generalizability in data scarce settings. Meanwhile, exploring the use of physics informed machine learning, which incorporates physical constraints into the learning process, may offer a promising direction to enhance model accuracy and robustness across diverse pathological conditions. Ultimately, these advancements could pave the way toward deployable, intelligent gait monitoring systems that support early diagnosis, personalized treatment planning, and continuous disease management in individuals with neurological disorders.

\section{Conclusion}
This study proposed a deep learning framework based on a CNN-BiLSTM architecture to estimate bilateral vGRFs in PD patients using wearable IMU data. This model achieved high accuracy ($R^2 \geq 0.91$) in both intra-subject and inter-subject scenarios between HC and PD patients separately, demonstrating robustness across varying gait conditions and subject-specific variability. Significant differences in sensor configurations were observed between PD patients and HC. Furthermore, we systematically assessed the influence of sensor placement and quantity, identifying optimal configurations that balance estimation performance with wearability in PD patients. The best performance was achieved using four IMUs (LD, RA, RS, FH), while a reduced configuration with only two IMUs (LD and FH) achieved comparable accuracy, suggesting strong potential for real-world deployment. This study firstly proposes a deep learning based framework for estimating bilateral vGRFs in PD-specific pathological gait using a reduced and optimized set of wearable IMUs. These findings offer a scalable solution for clinical gait assessment, remote monitoring, and personalized rehabilitation in neurological populations. Overall, this work lays the groundwork for the development of wearable and intelligent gait analysis systems tailored to pathological movement disorders.

\section*{Acknowledgment}
For the purpose of open access, the author has applied a creative commons attribution (CC BY) licence to any author accepted manuscript version arising.

\section*{References}
\bibliographystyle{IEEEtran} 
\bibliography{ref} 

\end{document}